\documentclass[conference,letterpaper,twoside,final,10pt]{ieeeconf}

\usepackage{graphicx}
\usepackage{amssymb}
\usepackage{amsmath}
\usepackage{cite}
\usepackage{comment}
\usepackage{color}
\usepackage{url}
\usepackage{alltt}
\usepackage{cleveref}
\usepackage{tikz}
\usepackage{pgfplots}
\usepackage{pgf-umlsd}
\usepackage{cprotect}
\usepackage{multirow}
\usepackage{algorithm}
\usepackage[noend]{algpseudocode}
\usepackage{bigfoot}
\usepackage{paralist}
\usepackage{tabularx}
\usepackage[whole]{bxcjkjatype}
\usepackage{ifthen}
\usepackage{threeparttable}
\usepackage[hang,small,bf]{caption}
\usepackage[subrefformat=parens]{subcaption}
\usepackage[super]{nth}
\usepackage{colortbl}
\usepackage{scalefnt}
\usepackage{threeparttable}

\pgfplotsset{compat=1.14}
\IEEEoverridecommandlockouts

\title{\LARGE \bf
Invisible Marker:\\Automatic Annotation of Segmentation Masks for Object Manipulation
}

\author{
Kuniyuki Takahashi$^{*\dagger}$,
Kenta Yonekura$^{*\dagger}$
\thanks{$^{*}$ The starred authors are contributed equally.}
\thanks{$^{\dagger}$ K. Takahashi and K. Yonekura are associated with Preferred Networks, Inc.
\ takahashi@preferred.jp \ yoneken@preferred.jp}}

\begin{document}

\maketitle
\thispagestyle{empty}

\begin{abstract}
We propose a method to annotate segmentation masks accurately and automatically using \emph{invisible marker} for object manipulation.
\emph{Invisible marker} is invisible under visible (regular) light conditions, but becomes visible under invisible light, such as ultraviolet (UV) light.
By painting objects with the \emph{invisible marker}, and by capturing images while alternately switching between regular and UV light at high speed, massive annotated datasets are created quickly and inexpensively.
We show a comparison between our proposed method and manual annotations.
We demonstrate semantic segmentation for deformable objects including clothes, liquids, and powders under controlled environmental light conditions.
In addition, we show demonstrations of liquid pouring tasks under uncontrolled environmental light conditions in complex environments such as inside the office, house, and outdoors.
Furthermore, it is possible to capture data while the camera is in motion so it becomes easier to capture large datasets, as shown in our demonstration.~\footnote{An accompanying video is available at the following link:\\ \url{https://youtu.be/fnpyDYUvDA4}}~\footnote{Dataset is available at the following link:\\ \url{https://github.com/pfnet-research/Invisible_marker_IROS2020}}
\end{abstract}
\section{Introduction}
\label{sec:introduction}
Accurate object recognition is one of the important functions for robots in order to manipulate objects.
In particular, dealing with deformable objects such as clothes, liquids, and powders is challenging, but it is important for a number of tasks such as laundry tasks and applications in biology and the medical area, which could benefit from robots that do laboratory experiments fully automatically.
However, these deformable objects are challenging to recognize accurately.

Deep learning has succeeded in computer vision (CV), natural language processing (NLP) and in the robotics area~\cite{long2015fully, ronneberger2015u, badrinarayanan2017segnet, young2018recent, hatori2018interactive, takahashi2019deep}, but generally requires massive datasets for training to achieve good performance.
Even though massive datasets are required, creating datasets consumes enormous costs such as money, time, and human resources.
Many objects are still beyond the reach of deep learning recognition because of the difficulty of creating large datasets.

\begin{figure}[tb]
	\centering
	\includegraphics[width=0.75\columnwidth]{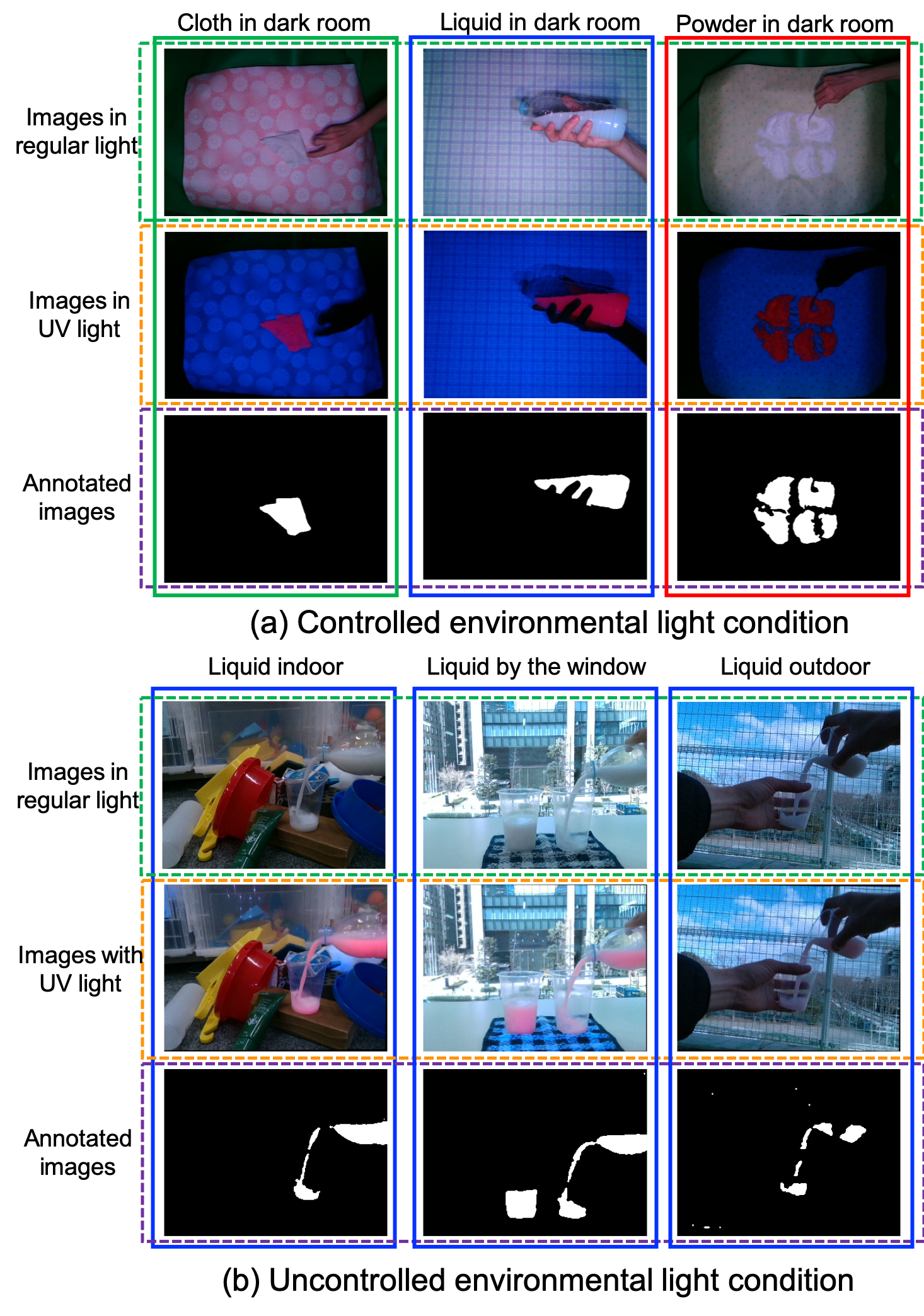}
	\caption{Dataset creation using \emph{invisible marker} for three kinds of objects: clothes, liquid, and powder.
	There are two conditions: environmental light is (a) controlled or (b) uncontrolled.
	}
	\label{fig:annotation}
	\vspace{-2mm}
\end{figure}

In this study, we focus on automatic annotation of segmentation masks for object manipulation to create large and accurate datasets quickly and inexpensively.
We use a marker that we call \emph{invisible marker}, which emits light when light outside the visible spectrum (invisible) is applied, as opposed to visible light, such as ultraviolet (UV) light.
This marker does not change the appearance of objects under visible (regular) light (See Fig.~\ref{fig:annotation}).
We show a comparison between manual annotations and our proposed method, and show its effectiveness for automatic annotation of deformable objects including clothes, liquids, and powders under controlled environmental light conditions (e.g. inside a dark room).
Additionally we show demonstrations of pouring liquid under uncontrolled environmental light conditions (referred to as \textbf{regular light conditions}) such as in offices, houses, and outdoor conditions even during daylight.

The rest of this paper is organized as follows.
Related works is described in~\ref{sec:related works}, and our contributions are explained in~\ref{sec:contributions}, while section~\ref{sec:invisible marker} details our proposed method.
Section~\ref{sec:experiments_cotrolled} outlines our experiment setup and evaluation settings under controlled light conditions, with results presented in~\ref{sec:results_controlled}.
Additionally, the same things under regular light conditions are described in sections~\ref{sec:experiments_noncotrolled} and~\ref{sec:results_noncontrolled} respectively.
Section~\ref{sec:discussion} discusses the cost of the proposed method.
Finally, conclusions are described as in section~\ref{sec:conclusion}.

\section{Related Works \& Contributions}
\label{sec:related works}
Annotation methods have been being developed to reduce the effort of creating datasets.
These methods fall in the two categories, manual annotation and automatic annotation.

\subsection{Manual Annotation}
Popular semantic segmentation datasets, such as Pascal VOC, MS-COCO, and COCO-Stuff, are manually annotated~\cite{everingham2010pascal, lin2014microsoft, caesar2018coco}.
Reduction of the workers' manual efforts by annotation tools~\cite{russell2008labelme, apolloscape_arXiv_2018, 2018arXiv180504687Y, andriluka2018fluid} and using crowdsourcing~\cite{hatori2018interactive, chang2017matterport3d} enables the creation of large-scale datasets.
One of the challenges of manual annotation is that it is prone to human errors and individual judgments in ambiguous cases.

\subsection{Automatic Annotation}
To prevent human errors and to create accurate datasets easily, automatic annotation methods have been developed.
The studies can be roughly classified into four groups. 

1) Approaches that focus on the features of the object itself, such as color tracking~\cite{liensberger2009color}, object temperature measurement (for example for hot liquids) using thermography~\cite{schenck2016detection}, and movement through background subtraction~\cite{brutzer2011evaluation}:
These approaches fail in annotating objects when a feature of an object is shared among multiple objects or the environment.

2) Approaches that focus on features provided in advance such as augmented reality (AR) markers~\cite{brachmann2014learning, kehl2017ssd}:
Even if there are multiple objects, the objects can be distinguished by different markers on each object. 
However, if a feature like a marker is attached to the object, it doesn't look the same anymore as without the marker.

3) Simulation approaches~\cite{hong2018virtual, aleksi2019affordance}:
This method can artificially create large datasets by switching between various background images and by capturing images of the target object from multiple perspectives.
However, the quality of the dataset is usually low for objects that are difficult to simulate, such as deformable objects.
The gap between simulation and the real world is also a challenge that needs to be solved, for example through sim-to-real transfer learning.

4) Approaches that focus on in advance provided markers which can be detected by specific devices, or by controlling external factors, for example by using special light.
The appearance of the object is not changed greatly under normal conditions.
Only the part of the object on which, for example, fluorescent paint has been applied will become visible under UV light.
In the medical field, a fluorescent substance that is attached to the material can be observed using a fluorescence microscope~\cite{dufour2005segmenting, rizk2014segmentation}.
In the computer vision field, a method that involves applying fluorescent paint to objects has been studied~\cite{baker2011database}.
To create optical flow datasets, a computer repeatedly takes a pair of images under both regular light and under UV light, and then moves the scene or camera by a small amount~\cite{baker2011database}.

Our proposed method belongs to the 4th category, which can be used to automatically create a segmentation mask as annotation using \emph{invisible marker} for deep learning.

\section{Contributions}
\label{sec:contributions}
Our main contribution is to expand previous work~\cite{baker2011database} as a general data creation method and to create a large dataset.
The details of our contributions are as follows.

1) Application of the method to deformable object manipulation (Section~\ref{sec:Annotation result by invisible marker}):
In previous research~\cite{dufour2005segmenting, rizk2014segmentation, baker2011database}, the target objects and materials have clear contours and are rigid and stationary.
In our study, rigid objects can be handled by our proposed method, but deformable objects were the main target.
Additionally, we collected datasets of deformable objects during manipulation since the objects are deformed during motions.
The challenges that arise during deformable object manipulation and their solutions were described.

2) Comparison with manual annotations by people and our proposed method to evaluate accuracy (Section~\ref{sec:Comparison with Proposed Method and Manual Method}): 
The dataset created with fluorescent paint is treated as ground truth for evaluating other methods, but the accuracy of the method using fluorescent paint has not been evaluated~\cite{baker2011database}.
We compared our proposed method and manual annotations by people using crowdsourcing.

3) Training a deep neural network on a dataset of deformable objects (Section~\ref{sec:Results of Semantic Segmentation}):
The previous study using fluorescent paint~\cite{baker2011database} was meant for optical flow, thus there was no need for a large dataset.
The datasets were too small to apply deep learning, despite the potential to create large datasets.
We have collected enough data for training and have verified that the generalization performance of deep learning can cope with deformable objects.

4) Investigation of the appearance changes caused by the fluorescent paint (Section~\ref{sec:Inference for Non-mixed Fluorescent Object}):
In~\cite{dufour2005segmenting, rizk2014segmentation,baker2011database}, the appearance change of an object by applying fluorescent paint was out of the scope of their work, because appearance changes do not influence the results in these works.
In our study, however, if appearance changes by fluorescent paint are large, the inference performance for semantic segmentation will decrease, particularly for the objects not painted with fluorescent paint.
We investigated whether the generalization capabilities of deep learning can still successfully deal with objects to which no fluorescent paint has been applied.

5) Application of several colors of fluorescent paints to multiple objects (Section~\ref{sec:Extension of proposed method}):
We show that the annotation for instance segmentation can be done by applying different colors of fluorescent paint in to multiple objects in a scene.

6) Proposal of a method to create datasets indoors as well as outdoors during daylight when light conditions cannot be controlled (Section~\ref{sec:Annotation result in Uncontrolled Light}):
In the previous study~\cite{baker2011database}, datasets could only be collected in a dark room where external light could be controlled.
Thus, collecting datasets was only possible in small experiment environments.
In this study, we proposed a method for creating datasets indoors as well as outdoors where external light cannot be controlled.

7) Proposal of a method to collect datasets while camera is in motion (Section~\ref{sec:Result of Image Transformation For Unfixed Camera Viewpoints}):
Although using a camera in a fixed position does not affect the quality of the annotation, it does limit the data collection process.
Moving a camera around, however, makes it difficult to collect pairs of images taken subsequently, though it would allow us to easily and quickly create a diverse dataset from different viewpoints.
We proposed a method that absorbs the camera movement even if the camera moves when capturing images.

8) Creation of a large dataset (Section~\ref{sec:Data Collection} \& \ref{sec:Data Collection For Uncontrolled Light}):
Our proposed method can create a dataset easily at high speeds and low costs.
32845 datasets are created in our study.

\section{Invisible Marker}
\label{sec:invisible marker}
\emph{Invisible marker} emits light under light outside the visible spectrum, and emits weaker light or none under visible (referred to as regular) light.
By using material that is (near) transparent under regular light for the \emph{invisible marker}, the appearance of objects under regular light do not change when \emph{invisible marker} is applied to them; any material can be used as an \emph{invisible marker} as long as it satisfies this condition.
Depending on the target object, sprays or liquids can be used as an \emph{invisible marker}.
We use fluorescent paint since it can be procured easily.
The object painted with fluorescent paint is luminescent under ultraviolet (UV) light; this becomes particularly evident in the dark, but is still visible when UV light is applied even in regular light.
Thus, our method can be used under both (1) \textbf{controlled} environmental light conditions (referred to as dark conditions) and (2) \textbf{uncontrolled} environmental light conditions (referred to as regular light conditions).

1) Objects without fluorescent paint are not visible in the dark even with a UV light turned on, but painted objects are visible.
When an image is captured under this UV light condition, only the part painted with the fluorescent paint becomes visible (See second-row in Fig.~\ref{fig:annotation}).
Only the painted part will be annotated as a class label for segmentation, by applying a threshold value on the RGB values of the image.

2) Under regular light, everything in the scene is still visible but the part to which fluorescent paint has been applied changes in appearance under UV light.
Given a pair of images captured in regular light conditions, one with and one without applying UV light, the region where only fluorescent paint was applied to can be obtained by calculating the difference between these images.
Then, in the same way as in dark conditions, a segmentation mask can be obtained by applying a threshold to the RGB values.

When moving the camera and taking pictures simultaneously, while also alternately switching between regular and UV light, the camera may move before capturing the same scene under different lighting conditions.
Thus, a shift of viewpoints between the captured images with and without UV light will occur.
If a segmentation mask is then created using the above method, the annotation mask will no longer match as a paired image captured under regular light because of this shift (The experiment details will be presented in section~\ref{sec:Result of Image Transformation For Unfixed Camera Viewpoints}).
In particular, when the camera is moved under regular conditions (point (2) from the previous paragraph), this shift will be a problem since a segmentation mask is calculated from the difference in a captured pair of images with and without applying UV light.
Therefore, it is necessary to transform the images so that the viewpoints of the two images match.
The transformation process is done in three steps: 1) The image features from the images captured in regular and UV light conditions are extracted, 2) the feature point of image captured under regular light corresponding to another feature point of image captured under UV light with the shortest distance is matched, and 3) the image captured under UV light is transformed into a form that fits the image captured under regular light.
The feature extraction method uses Oriented FAST and Rotated BRIEF (ORB), because of the calculation cost, matching accuracy~\cite{rublee2011orb}.
A simple brute-force matcher is used for feature matching.
For image transformations, we used homography transformations, which can be used to transform rectangles into trapezoids, unlike affine transformations, by projecting a plane onto another plane through projective transformations.

\begin{figure}[tb]
    \vspace{2mm}
	\centering
	\includegraphics[width=0.6\columnwidth]{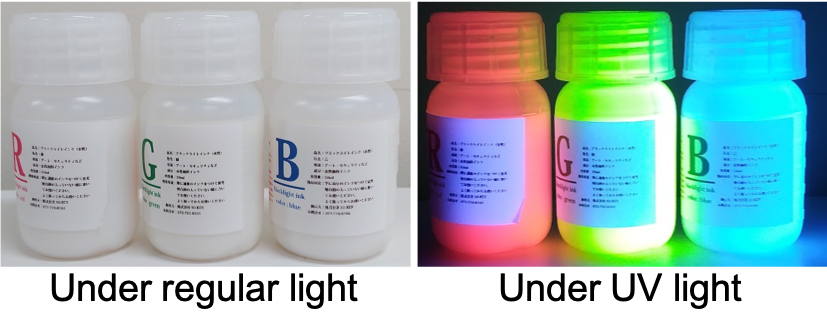}
	\caption{Fluorescent paint under regular light and UV light}
	\label{fig:fluorescent_paint}
	\vspace{-2mm}
\end{figure}

Since only the target object can be extracted by \emph{invisible marker}, our proposed method does not depend on the complexity of the background, or anything not painted with fluorescent paint, and only the fluorescent painted objects are extracted.
Moreover, our method can distinguish individual objects from multiple objects by applying different colors to each object since various colors of fluorescent paints can be created by mixing red, green, and blue color (Fig.~\ref{fig:fluorescent_paint}), which is useful for instance segmentation.
More details will be discussed in Section~\ref{sec:Extension of proposed method}.

\section{Experiment Setup For Controlled Light}
\label{sec:experiments_cotrolled}
The purpose of the experiments under controlled environmental light conditions (dark conditions) is to 1) verify the accuracy of the datasets for deformable objects created with the \emph{invisible marker} (Section~\ref{sec:Annotation result by invisible marker}), 2) to compare the proposed method with manual annotations by people (Section~\ref{sec:Comparison with Proposed Method and Manual Method}), 3) to evaluate the created dataset for deep learning (Section~\ref{sec:Results of Semantic Segmentation}), 4) to investigate the effect of appearance changes by \emph{invisible markers} (Section~\ref{sec:Inference for Non-mixed Fluorescent Object}), and 5) to apply several colors of fluorescent paints for multiple objects in one scene and observe the results (Section~\ref{sec:Extension of proposed method}).

\begin{figure}[tb]
	\centering
    \includegraphics[width=0.9\columnwidth]{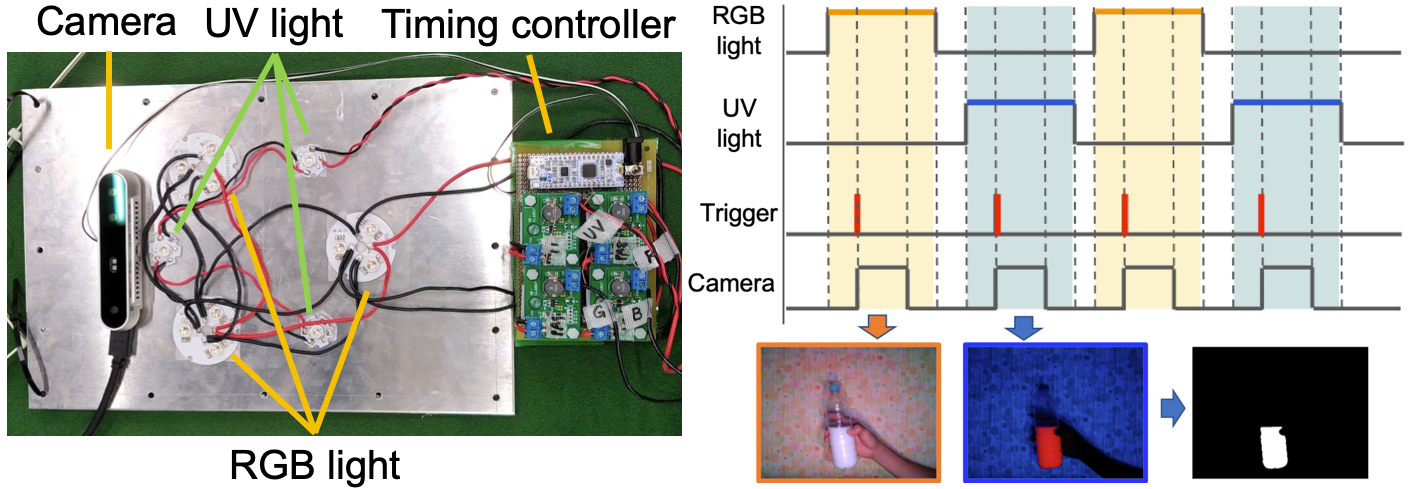}
    \caption{The whole capturing system and timing chart.}
    \label{fig:system}
\end{figure}

\subsection{Data Acquisition Device}
\label{sec:Data acquisition device}
Datasets for segmentation masks require a set of images under regular light as input data paired with output data that contains a label that tells to which class each pixel of the image belongs.
We developed a system that can capture images under the regular light and invisible (UV) light to create such paired images automatically using \emph{invisible marker} (See Fig.~\ref{fig:system}).
Our system is composed of three parts:
\begin{itemize}
    \item Camera part which captures images of a target object
    \item Lighting part which controls the lighting output
    \item Control part which controls the timing of capturing images and changing the light conditions
\end{itemize}

For the camera part, we use a camera for which the capture timing can be controlled through an external trigger input.
When the camera receives an external trigger input from the control part, the camera captures an image of the target object.
Then, the image is sent to the control computer.
In this experiment, we use a RealSense D415 camera~\cite{keselman2017intel}.

For lighting, we use an RGB LEDs as regular light, and UV LEDs as ``invisible'' light.
Power LED drivers are used to control them.

For the control part, a NUCLEO-F303K8 board~\cite{nucleo} provides trigger signals to the lighting part to control the emission timings and intensities.
The NUCLEO-F303K8 board can be controlled from the computer through USB.
At the same time, the control part also outputs a trigger signal to the camera part to control the capture timing (See timing chart in Fig.~\ref{fig:system}).
To create datasets of segmentation masks for dynamically changing objects, regular lights and UV lights are alternately switched at high speed.

\subsection{Target objects}
Accurate recognition of deformable objects such as clothes, liquids, and powders is important in the industry and daily life to e.g. fold clothes, manipulate medicine, and cook.
We prepared a cloth (handkerchief), liquid (water) and powder (baking soda) and mixed them with fluorescent paint.

\subsection{Data Collection}
\label{sec:Data Collection}
\begin{figure}[tb]
    \vspace{2mm}
	\centering
	\includegraphics[width=1.0\columnwidth]{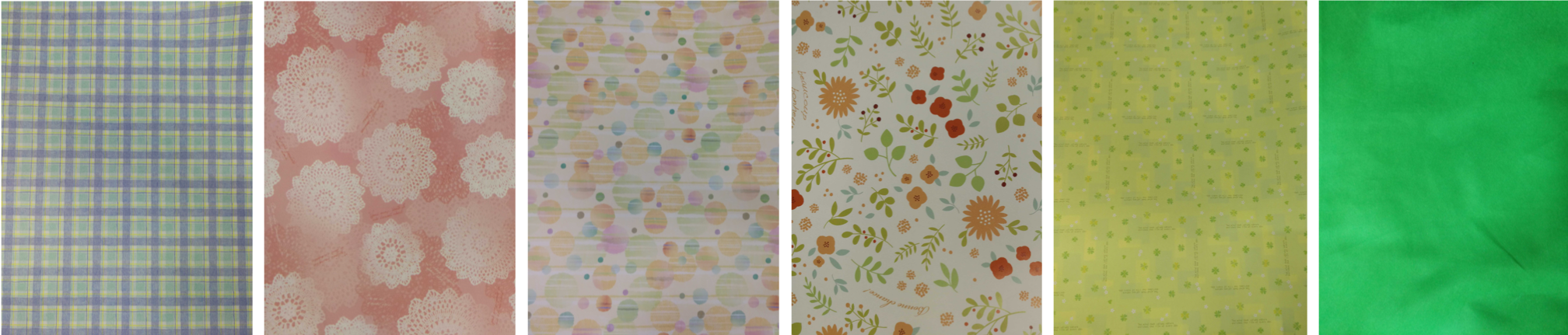}
	\caption{Six type of backgrounds for data collection}
	\label{fig:background}
	\vspace{-3mm}
\end{figure}

For data collection, images are captured by the data acquisition system described in Section~\ref{sec:Data acquisition device} during motions; folding of a cloth, shaking liquid in a plastic bottle, and stirring powder with a spoon.
The sampling rate of the camera is 30 Hz.
The sampling rate for creating a paired image dataset is 15 Hz because images are taken by alternately switching between regular light and UV light.
We prepared six different types of backgrounds (Fig.~\ref{fig:background}) and perform one motion per background, thus a total of six motions per object.
The total number of acquired images for clothes, liquids, and powders are 3920, 3081, and 4199, respectively.
The dataset size is sufficient to achieve high accuracy for inference by multiple deep learning models.
Details will be described in section~\ref{sec:Results of Semantic Segmentation}.
Datasets of five out of six backgrounds are used for training and the data of one remaining background is used for evaluation through 6-fold cross-validation as all combinations of backgrounds.
This means that evaluation is performed on untrained background.
Acquired images are resized from $640\times480$ to $160\times120$.

\subsection{Deep Learning \& Training}
\label{sec:Deep Learning}
In order to show the effectiveness of the created datasets using \emph{invisible marker}, we verify that they can be used to train three typical kinds of deep learning models for semantic segmentation: FCN~\cite{long2015fully}, U-Net~\cite{ronneberger2015u}, and SegNet~\cite{badrinarayanan2017segnet}.
These models are trained with datasets composed of images in regular light as input and the annotated data created by the proposed method as output.
Chainer, a deep learning library, is used for the implementation~\cite{tokui2015chainer}.
All our network experiments were performed on a machine equipped with 256\,GB RAM, an Intel Xeon E5-2667v4 CPU, and eight Tesla P100-PCIE with 12GB.
Training time for each material was within 30 minutes by parallel processing with 8 GPUs.

\section{Experiment Results in Controlled Light}
\label{sec:results_controlled}
\subsection{Annotation result by invisible marker}
\label{sec:Annotation result by invisible marker}
We first show examples of created datasets for deformable object manipulation (Fig.~\ref{fig:annotation} (a)).
In Fig.~\ref{fig:annotation} (a), we can observe that fine unevenness at the edges of the powder is annotated accurately.
In addition, only the target object is annotated even though the background includes similar colors to the objects, and other objects such as a hand and a spoon are ignored since they are not painted with fluorescent paint.
Results for more complex backgrounds will be described in section~\ref{sec:results_noncontrolled}.
Conventional methods, such as focusing on colors or background subtraction, usually have difficulties to annotate in these situations.
Methods that focus on color cannot annotate correctly if the background color is similar to the target object.
In the background subtraction method, not only the target object but also the hand, spoon, and plastic bottle will be annotated.
Furthermore, these methods have difficulties in handling multiple objects and instance segmentation, as will be described in Section~\ref{sec:Extension of proposed method}.


\subsection{Comparison b/w Proposed Method and Manual Method}
\label{sec:Comparison with Proposed Method and Manual Method}
The purpose of this section is to verify the accuracy of our method.
We compare manual annotations to our created annotations created by using the \emph{invisible marker}.
In order to strictly compare the proposed method with manual annotations, it is necessary to create a manual annotation dataset and use it to train a network.
However, even though the number of images in our dataset is only 11200, this experiment is difficult due to budgeting reasons as the manual annotation cost would be around \$ 10000, thus we were unfortunately unable to conduct this experiment.

As an alternative to this costly method, two comparisons from a small sample were performed by using the intersection over union (IoU), which is simply a ratio of the union of two regions and the intersection of the two regions.
The two comparisons are as follows:
\begin{enumerate}
    \item Apply IoU only between manual annotations to measure individual differences, and
    \item Apply IoU to the annotations created by the proposed method and the manual annotations to measure how close the proposed method is to manual annotations.
\end{enumerate}
If both IoU values are similar, it means that our method using \emph{invisible marker} is at least as accurate as human annotations.
As for the comparisons, we selected three images for each object and each background, that is totally 54 images, and three people per image are assigned for annotation using Amazon Mechanical Turk (AMT).

\subsubsection{Differences Between Manual Annotations}
\begin{figure}[tb]
	\centering
	\includegraphics[width=0.75\columnwidth]{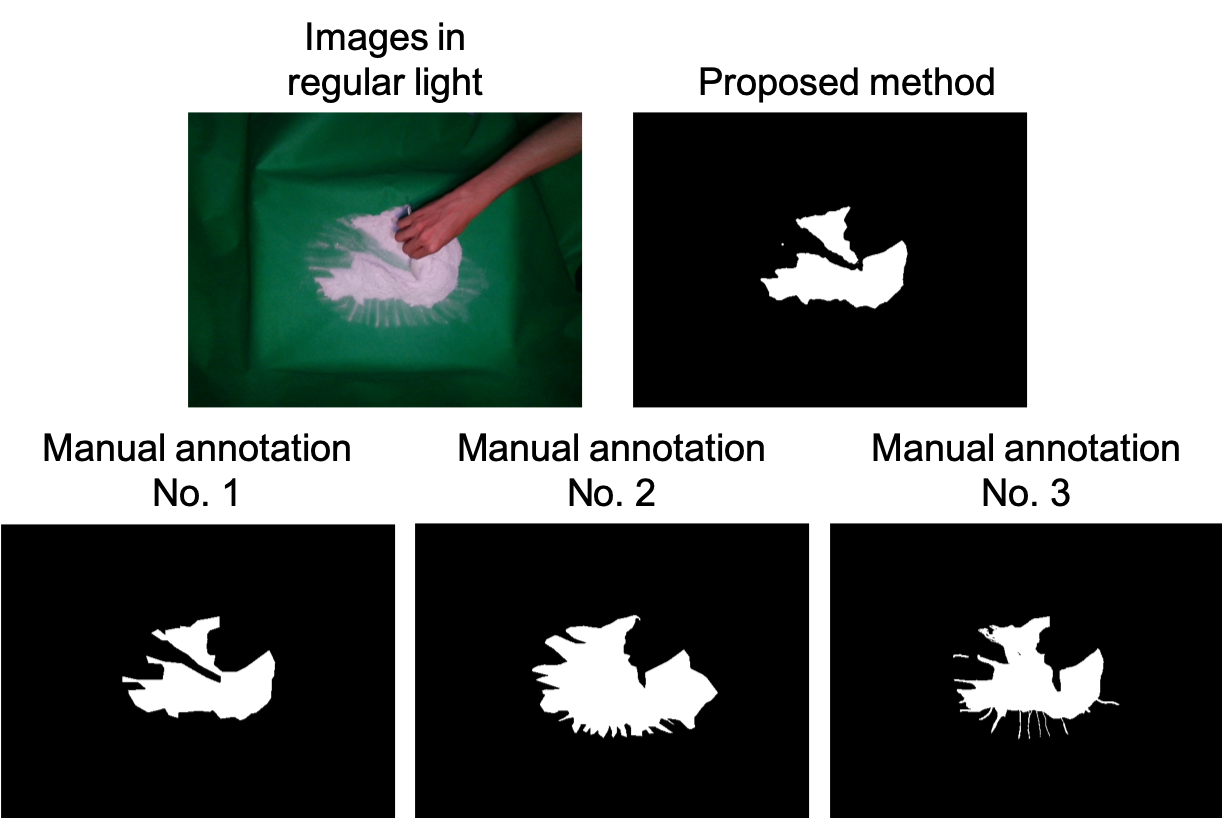}
	\caption{Annotated images for powder by proposed method and manual by three people. 
	}
	\label{fig:powder_annotation}
	\vspace{-1mm}
\end{figure}

Table~\ref{tab:comparison} shows the mean value and standard deviation about the comparisons.
In Table~\ref{tab:comparison}, the IoU among manual annotations shows that individual differences seem to depend on the clarity of boundaries between the target object and other objects.
Differences among different annotators are small for clothes and liquids because these boundaries are clear.
On the other hand, individual differences are large when there is an unclear boundary between the object and others like small amount of powder around the boundaries. 
This can also be seen from Fig.~\ref{fig:powder_annotation}, which shows images of both manually and automatically annotated powder.
It can be seen that the manual annotations are clearly different from each other.

Manual annotator No.1 ignores fine powder area, while manual annotator No.2 tries to cover the entire area of the fine powder, and manual annotator No.3 is a type in between No.1 and No.2.
When people annotated ambiguous scenes such as with powder, individual judgments creates variety in the datasets.
Our proposed method can control the annotation result by adjusting the amount of fluorescent paint to control the light intensity emitted under UV light, and the threshold for the RGB values for extracting the segmentation mask.
In this experiment, the amount of fluorescent paint and the threshold value are adjusted to ignore fine powder like manual annotator No. 1 does, since this area is generally too fine to be manipulated by robots.

\begin{table}[tb]
    \vspace{2mm}
    \centering
    \caption{Comparison of annotations between the manual method and the proposed method. }
    \label{tab:comparison}
    \begingroup
    \scalefont{0.85}
        \begin{tabular}{c|ccc}
        \hline
         & cloth & liquid & powder\\
        \hline\hline
         	\begin{tabular}{c} IoU among manual \end{tabular}&
         	\begin{tabular}{c} 94.0\%\\$(\pm{3.55\%})$\end{tabular}&
         	\begin{tabular}{c} 93.6\%\\$(\pm{2.21\%})$\end{tabular}&
         	\begin{tabular}{c} 84.6\%\\$(\pm{10.4\%})$\end{tabular}\\
         	
            \begin{tabular}{c}IoU b/w manual \& \\ \emph{invisible marker}\end{tabular}&
            \begin{tabular}{c} 89.8\%\\$(\pm{7.19\%})$\end{tabular}&
            \begin{tabular}{c} 77.6\%\\$(\pm{9.25\%})$\end{tabular}&
            \begin{tabular}{c} 84.0\%\\$(\pm{8.95\%})$\end{tabular}\\
        \hline
        \end{tabular}
    \endgroup
    \vspace{-3mm}
\end{table}
\begin{figure}[tb]
    \vspace{2mm}
	\centering
	\includegraphics[width=0.70\columnwidth]{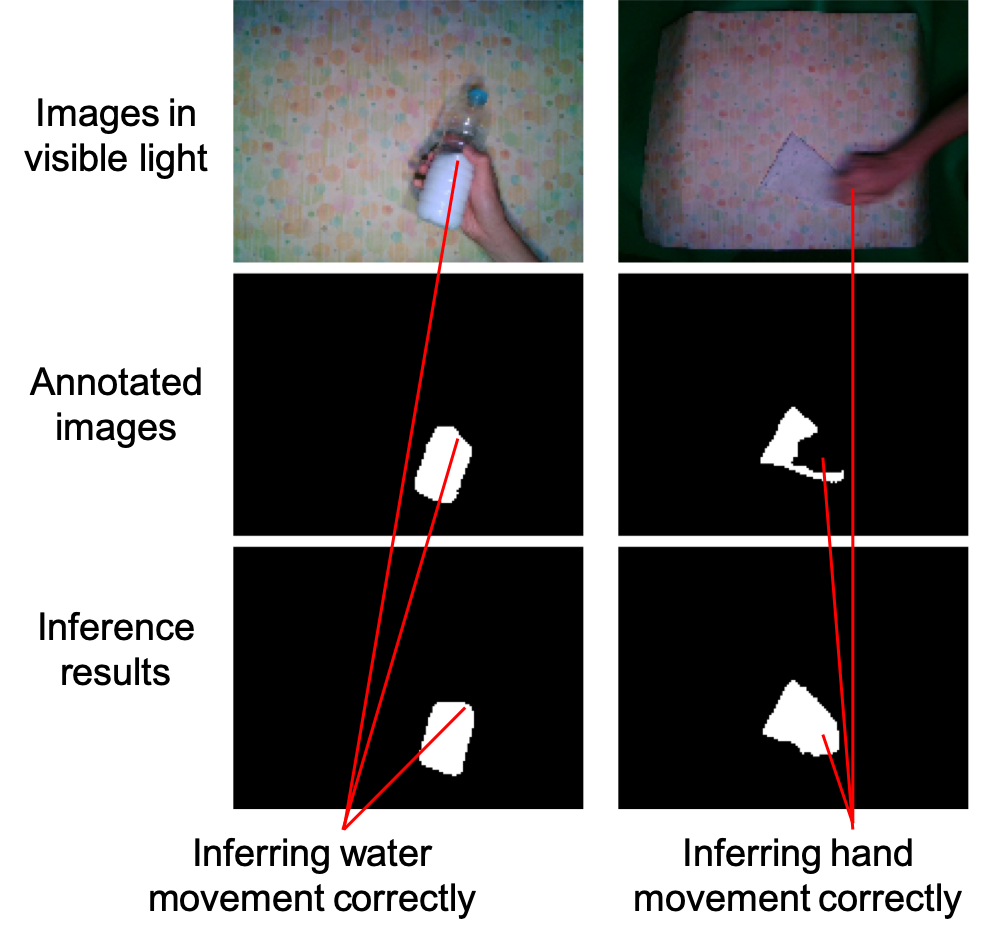}
	\caption{Images for annotation gaps due to the fast movements and the gaps absorbed by deep learning}
	\label{fig:annotation_gap}
\end{figure}
\begin{table}[t]
    \centering
    \caption{Comparison of annotations between the manual and proposed method using stationary liquid}
    \label{tab:comparison_static}
    \begin{tabular}{c|c}
    \hline
     & liquid\\
    \hline\hline
     	IoU among manual                    & 94.6\% $(\pm{1.95\%})$ \\
        IoU b/w manual \& invisible marker  & 93.6\% $(\pm{1.41\%})$ \\
    \hline
    \end{tabular}
\end{table}
\subsubsection{Gaps b/w the Proposed Method and Manual Method}
\label{sec:gaps}
In Table~\ref{tab:comparison}, the result of IoU between manual annotations and \emph{invisible marker} shows that differences between manual annotation and \emph{invisible marker} are more significant in liquids than clothes and powders.
The first and second rows of Fig.~\ref{fig:annotation_gap} is one of the examples of a large gap between the image under regular light and the annotated images because of the fast movement of the objects and the hand.
The movement causes the liquid to shift, and the position of the human hand is also shifted on the cloth.
We think that the gaps for the liquid occurred more than for the cloth because the liquid moved more and faster than the cloth due to their inertial properties.

In order to investigate whether the capture timing is the main factor for causing the gaps, we evaluate the IoU in the same way as Table~\ref{tab:comparison} using stationary liquid on the desk instead of liquid in movement.
We selected four images of stationary liquid, and three people per image are assigned for annotation using AMT.
The result is shown in Table~\ref{tab:comparison_static}.
As a result, it can be seen that there is only a small difference between the proposed method and the manual method.
This camera hardware challenge can be alleviated if we use a high-speed camera, though there will always be a limitation to the camera sampling rate.
In this research, we focus on a software approach to deal with the gaps in the dataset.
By training a deep neural network on a large-scale dataset, such gaps are absorbed by the generalization capability of the network.
The experiment result will be described in section~\ref{sec:Results of Semantic Segmentation}.

\subsection{Results of Semantic Segmentation}
\label{sec:Results of Semantic Segmentation}
\begin{figure}[tb]
	\centering
	\includegraphics[width=0.70\columnwidth]{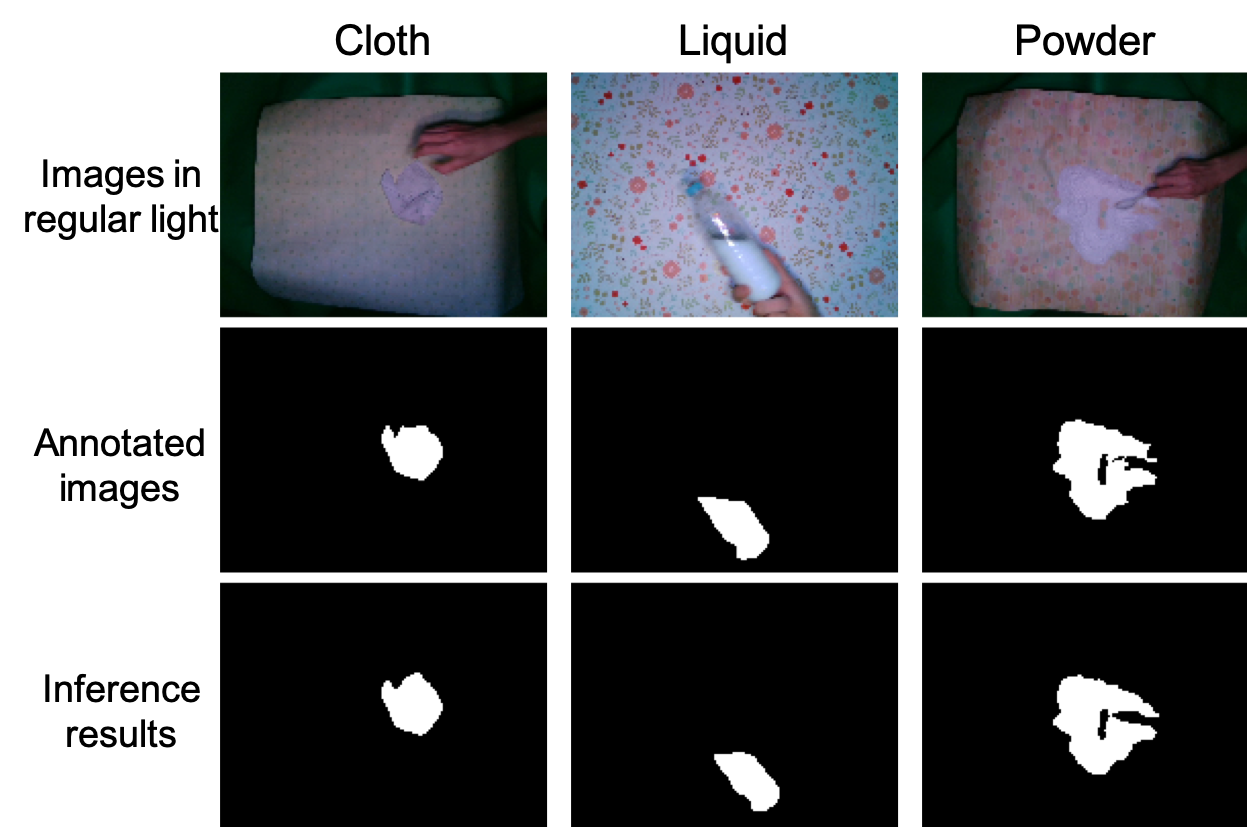}
	\caption{Inferred results of cloth, liquid, and powder}
	\label{fig:inference}
\end{figure}

The purpose of this section is to test whether deep learning can be trained with the invisible marker datasets, and to evaluate if deep learning has enough generalization capabilities to absorb the gap between the captured image under regular light and UV light caused by the fast movement of the objects as described in section~\ref{sec:gaps}.
Table~\ref{tab:accuracy} shows the IoU of the inferred segmentation mask on the validation data (which contains only untrained backgrounds), after applying 6-fold cross validation for 6 backgrounds.
In U-Net, the accuracy for validation data is over 80\% for all objects.
Fig.~\ref{fig:inference} shows the examples of inferred results of clothes, liquids, and powders on untrained background data.
As can be seen the accuracies are all around 80\% in Table~\ref{tab:accuracy} and from Fig.~\ref{fig:inference}, we can conclude that deep neural networks can be trained correctly with the created datasets.

As described in Section~\ref{sec:gaps}, the fast movement of the object creates a gap between captured images in regular light and UV light (See the first and second rows of Fig.~\ref{fig:annotation_gap}).
The inferred results in the third row of Fig.~\ref{fig:annotation_gap} shows that the generalization capability of deep learning absorbs these gaps, and thus can still accurately infer the segmentation masks.
These images are also inferred with untrained backgrounds.
We conclude that our proposed method can create automatic and accurate annotation datasets of deformable object manipulation and that the quality is sufficient to train deep neural networks to correctly infer segmentation masks.

\begin{table}[t]
    \vspace{2mm}
    \caption{IoU of semantic segmentation}
    \centering
    \begin{tabular}{c|ccc}
    \hline
     IoU & cloth & liquid & powder \\
    \hline\hline
     	FCN     & 76.7\% & 84.8\% & 83.1\% \\
        U-Net   & 81.3\% & 87.8\% & 87.7\% \\
        SegNet  & 72.7\% & 83.4\% & 84.4\% \\
    \hline
    \end{tabular}
    \label{tab:accuracy}
    \vspace{-2mm}
\end{table}
\subsection{Inference for Non-mixed Fluorescent Objects}
\label{sec:Inference for Non-mixed Fluorescent Object}
\begin{figure}[t]
    \vspace{2mm}
	\centering
	\includegraphics[width=0.65\columnwidth]{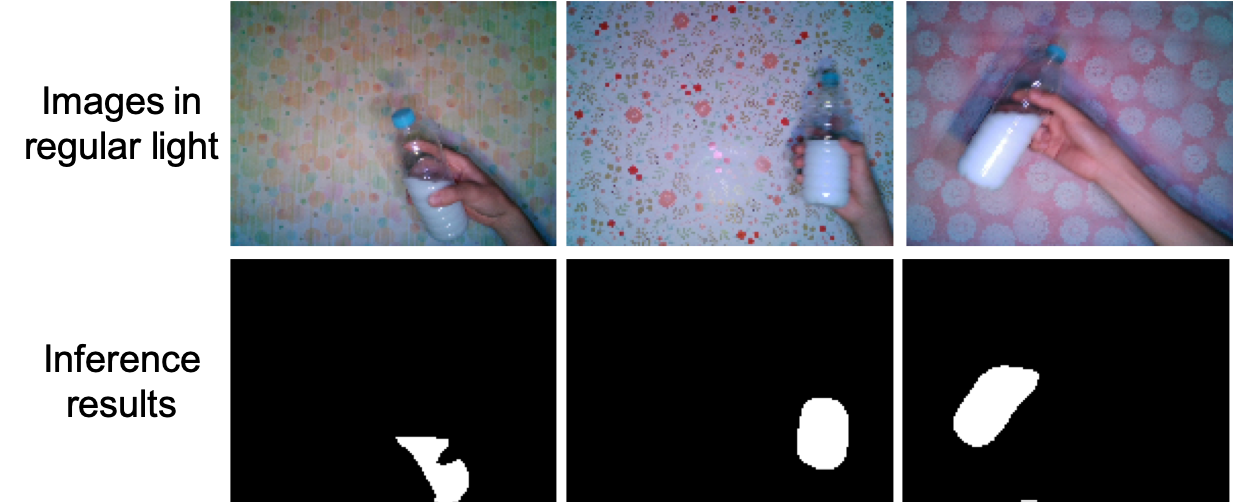}
	\caption{Inferred result of non-mixed fluorescent paint using the trained networks through the datasets with fluorescent paint}
	\label{fig:inference_wo_paint}
	\vspace{-2mm}
\end{figure}

To investigate the effects of appearance changes due to fluorescent paint, we trained the networks using the datasets created by our method and subsequently validated it on objects which are \textbf{not} mixed with fluorescent paint.
If the appearance is changed by the fluorescent paint, the network cannot infer the segmentation mask correctly for the object which is not mixed with fluorescent paint because it will simply look different from the training set.
Fig.~\ref{fig:inference_wo_paint} shows the inference result of the liquid.
The target object to be inferred is the liquid that would have had the greatest appearance change.
The inferred results show that the segmentation mask is still accurate despite the absence of the paint.
We can thus conclude that the change in appearance due to the fluorescent paint is small enough (if there is any) to be absorbed within the generalization capability of deep learning.

\subsection{Several Colors of Invisible Marker and Multiple Objects}
\label{sec:Extension of proposed method}
\begin{figure}[tb]
	\centering
	\includegraphics[width=0.60\columnwidth]{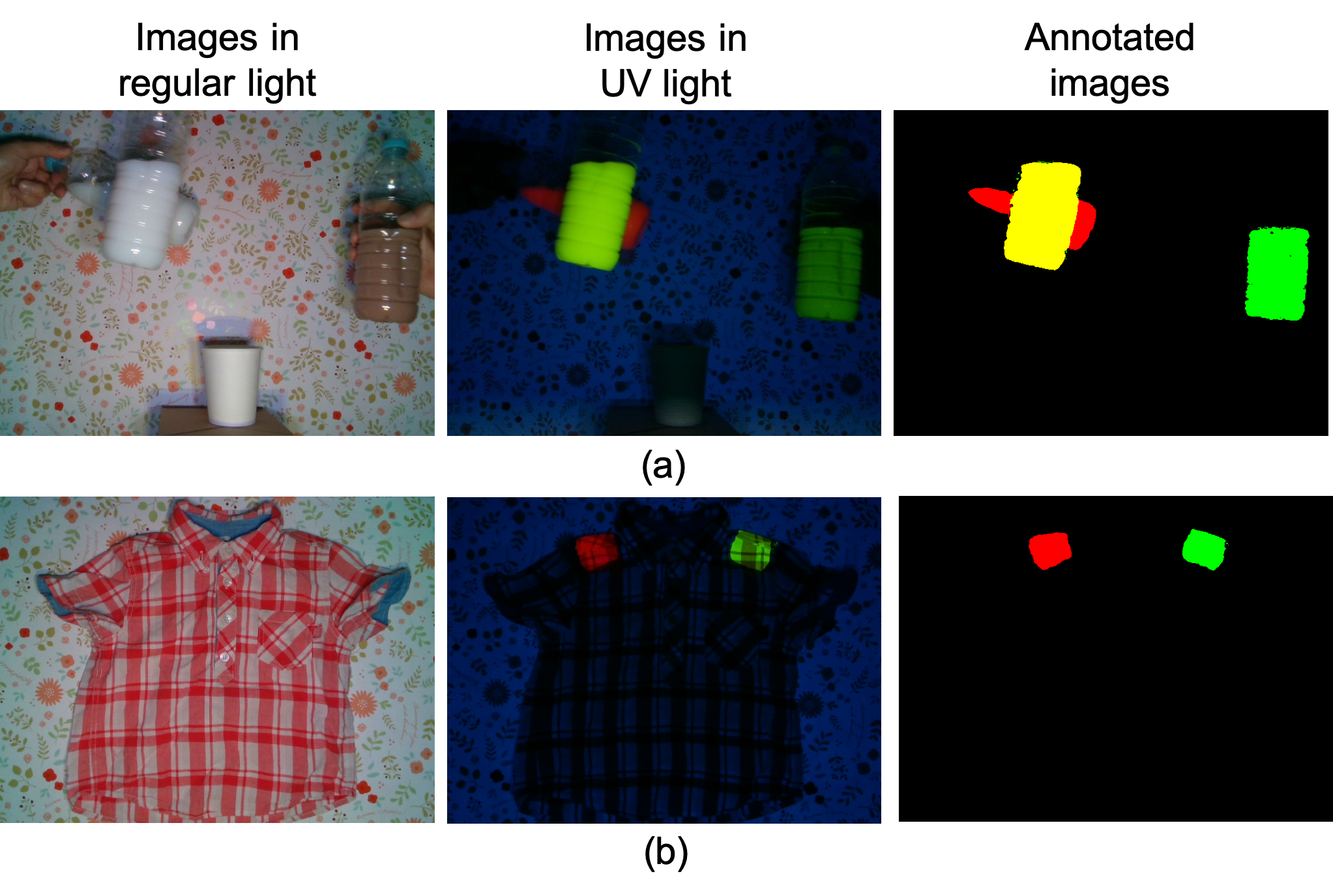}
	\caption{(a) Semantic segmentation for multiple object, overlapped situation, and instance segmentation (b) Grasping points}
	\label{fig:extension}
\end{figure}

In the proposed method, the color of \emph{invisible marker} can be changed for each object.
Even if there are multiple objects, each object can be painted with different colors.
Even in the situation where objects overlap, or same and/or multiple objects, each object can be annotated separately (Fig.~\ref{fig:extension} (a) shows the results of segmentation for multiple objects, overlapping objects, and instance segmentation of two white colored bottles and a brown colored bottle ).
In addition, it is also possible to create annotated datasets for only grasp points of an object by applying \emph{invisible marker} to only one part instead of the whole object.
Moreover, by giving different colors to each grasp point of the same object, robots can distinguish them from each other and manipulate them according to their purposes (Fig.~\ref{fig:extension} (b) shows the annotation of two grasping points using two different class labels).

\section{Experiment Setup Under Regular Light}
\label{sec:experiments_noncotrolled}
We conducted experiments under regular light conditions (uncontrolled environmental light conditions) similar to what we did under dark conditions (controlled environmental light conditions) described in section~\ref{sec:experiments_cotrolled}.

\subsection{Data Acquisition Device For Regular Light}
\label{sec:Data Acquisition Device For Uncontrolled Light Condition}
The configuration of the data acquisition device is the same except that we only blink a UV light and don't use regular lights anymore since there is enough external environmental light.

\subsection{Target object and Environment Under Regular Light}
\label{sec:Target object and Environment For uncontrolled Light}
Among the deformable objects, liquids are widely used indoors and outdoors, such as in medical drug factories and laboratories, and home environments.
We conducted liquid pouring tasks in various places e.g. indoors, by the window, and outdoors.
The liquid is mixed with fluorescent paint.

\subsection{Data Collection Under Regular Light}
\label{sec:Data Collection For Uncontrolled Light}
For data collection, images are captured during water pouring motions by the data acquisition system described in Section~\ref{sec:Data Acquisition Device For Uncontrolled Light Condition}.
The sampling rate of the camera is 30 Hz, thus the sampling rate for creating paired images is 15 Hz because images are taken while switching between UV and regular light.
The total number of acquired pairs are 21645.
Regular light conditions are difficult since the backgrounds are complicated and varied, so a large dataset was necessary.
Details will be described in section~\ref{sec:Annotation result in Uncontrolled Light}.
Acquired images are resized from $320\times240$ to $160\times120$.

\section{Results under Regular Light Conditions}
\label{sec:results_noncontrolled}
\subsection{Annotation result under Uncontrolled Light}
\label{sec:Annotation result in Uncontrolled Light}
We show examples of datasets for liquid pouring tasks captured under regular light conditions (uncontrolled environmental light conditions) (Fig.~\ref{fig:annotation}(b)).
In addition to the examples, experiments are performed at various places indoors, near windows, and outdoors, so please check the attached video file and our published data.
It is clear from Fig.~\ref{fig:annotation}(b) that the environmental light is not controlled, that is, it cannot be turned off while applying UV light; so not only the fluorescent paint but also everything else is visible.
When UV light is applied, it can be seen that only the liquid mixed with the fluorescent paint changes red.
It was expected that the fluorescent paint would always react and become red near the windows and outdoors since sunlight contains UV light. 
However, the UV light in sunlight is not strong enough to influence our experiment.
From the difference between the image under regular light and the one under UV light, an annotation for a segmentation mask was achieved by extracting only the part that emitted light because of the fluorescent paint.
From the above, it can be seen that the segmentation mask is created correctly even in a situation where the environmental light cannot be controlled and with realistic, complicated backgrounds.

\begin{figure}[tb]
    \vspace{2mm}
	\centering
	\includegraphics[width=0.95\columnwidth]{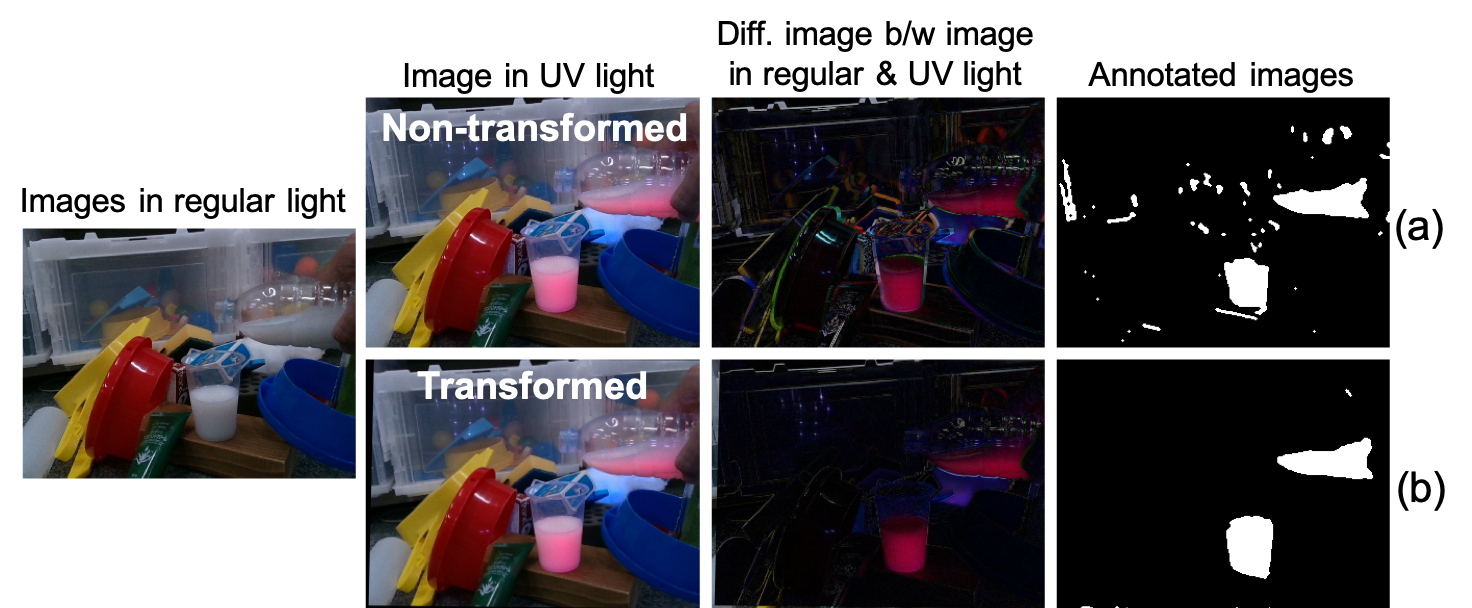}
	\caption{(a) without and (b) applying image transformation}
	\label{fig:image_transformation}
\end{figure}
\subsection{Result of Image Transformation For Unfixed Viewpoints}
\label{sec:Result of Image Transformation For Unfixed Camera Viewpoints}
In this section, we show the result of the image transformation to deal with differences of camera viewpoints between the image under regular light and the image under UV light, while moving the camera during capturing images.
When moving the camera while taking pictures simultaneously with alternately switching between regular and UV light, a shift of viewpoints between the captured images with and without UV light occurred.
Thus, the annotation mask no longer matches as a paired image captured under regular light because of this shift.
Therefore, image transformation is necessary for the viewpoints of the two images to match.
Fig.~\ref{fig:image_transformation} shows the comparison before and after applying the image transformation.
If the image transformation is not performed, not only the target liquid but also the background edges are annotated due to the viewpoint shift in images captured under regular light and UV light, respectively.
On the other hand, when the image transformation is performed, the influence of the shift due to the viewpoint transformation is reduced, and it can be seen that only the target object is annotated correctly.
From the above, it is shown that a dataset can be created without fixing the camera viewpoint, by performing this image transformation.
\subsection{Comparison b/w the Proposed and Manual Method}
\label{sec:Comparison b/w the Proposed Method and Manual Method}
In order to evaluate whether creating datasets under regular light conditions works like under dark conditions, we evaluate the IoU in the same way as we did for Table~\ref{tab:comparison_static}, only this time under regular light conditions.
We selected 9 images from our collected data in which the liquid is stationary.
If there is a gap between the proposed method and the manual method, we can conclude that regular light conditions do not affect our data creation process.
Three people per image are assigned for annotation using AMT, the comparison results are shown in Table~\ref{tab:comparison_static}.
As a result, it can be seen that there is only a small difference between the proposed method and the manual method even under regular light conditions.
Thus, we conclude that our method can create dataset accurately even under regular light conditions.

\begin{table}[t]
    \vspace{2mm}
    \centering
    \begingroup
    \scalefont{0.80}
    \caption{cf. annotations b/w the manual and proposed method using stationary liquid under regular light cond.}
    \label{tab:comparison_static_regular}
        \begin{tabular}{p{8.5em}|p{4.4em}p{4.8em}p{4.8em}|p{4.4em}}
    \hline
         & \begin{tabular}{c}Indoor\end{tabular} & \begin{tabular}{c}By window\end{tabular} & \begin{tabular}{c}Outdoor\end{tabular} & \begin{tabular}{c}Ave.\end{tabular}\\
        \hline\hline
         	\begin{tabular}{c} IoU among manual \end{tabular}&
         	\begin{tabular}{c} 85.2\%\\$(\pm{4.82\%})$\end{tabular}&
         	\begin{tabular}{c} 79.3\%\\$(\pm{7.65\%})$\end{tabular}&
         	\begin{tabular}{c} 92.1\%\\$(\pm{2.08\%})$\end{tabular}&
         	\begin{tabular}{c} 85.5\%\\$(\pm{7.40\%})$\end{tabular}\\
         	
            \begin{tabular}{c}IoU b/w manual \& \\ \emph{invisible marker}\end{tabular}&
            \begin{tabular}{c} 83.4\%\\$(\pm{2.56\%})$\end{tabular}&
            \begin{tabular}{c} 77.9\%\\$(\pm{6.42\%})$\end{tabular}&
            \begin{tabular}{c} 89.0\%\\$(\pm{3.66\%})$\end{tabular}&
            \begin{tabular}{c} 83.4\%\\$(\pm{6.33\%})$\end{tabular}\\
        \hline
    \end{tabular}
    \endgroup
    \vspace{-2mm}
\end{table}

\begin{figure}[t]
	\centering
	\includegraphics[width=0.75\columnwidth]{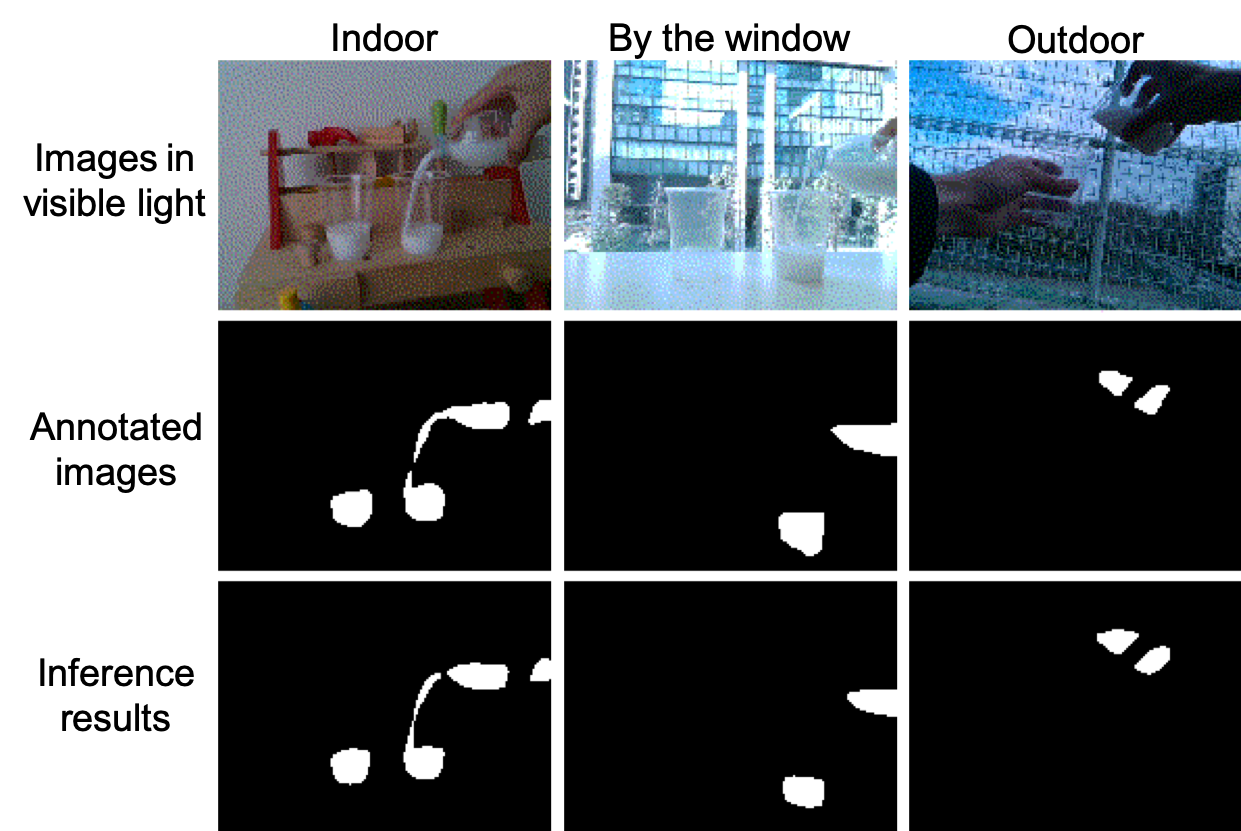}
	\caption{Inferred results of liquid pouring tasks}
	\label{fig:inferece_regular}
\end{figure}
\subsection{Results of Semantic Segmentation Under Regular Light}
\label{sec:Results of Semantic Segmentation Under Regular Light Cond}
The purpose of this section is to confirm whether deep neural networks can be trained with datasets acquired under regular light conditions.
The results of IoU of the inferred segmentation mask on the validation data are 87.1\%, 92.5\% and, 90.1\% for FCN, U-Net, and SegNet, respectively.
The networks can infer segmentation mask correctly in complex environments such as indoor, by the window, and outdoor conditions (See Fig.~\ref{fig:inferece_regular}. See more examples in the attached video).
We can conclude that deep neural networks can be trained correctly with the datasets created by our method.
\section{Discussion for Cost}
\label{sec:discussion}
In this section, we discuss the cost of time and money for manual annotations and the proposed method.
Table~\ref{tab:cost} shows the total cost for manual annotations and the proposed method.
Manual annotations are costly for time and money, whereas coloring objects and the data acquisition system (which is mostly a one-time cost) is cheaper.
In the proposed method, the application of the fluorescent paint took only a few minutes per object.
Time per image is negligible since painting time can be ignored considering that we only paint once but take a lot of images per object.
The cost of the fluorescent paint was \$ 18 for all objects for all experiments.
For the acquisition system, UV light as an additional device is required for the proposed method, but once we make the data acquisition system, it can be used again.
For manual annotations, the total cost increases as the number of images increases and the cost per image does not decrease as the datasets grow in size.
Even annotation tools such as LabelMe~\cite{russell2008labelme} can only reduce the time and cost per image, but the total costs will still increase as more images are annotated.
In the proposed method, cost of time and money per image decrease because colored objects can be reused once objects are painted by the \emph{invisible marker}.
Since deep learning requires thousands to tens of thousands of images, it is clear that the proposed method can create datasets quickly and inexpensively compared to manual annotations.

\begin{table}[tb]
    \vspace{2mm}
    \centering
    \begin{threeparttable}
    \caption{Cost for manual method and the proposed method}
    \label{tab:cost}
    \begingroup
    \scalefont{0.8}
        \begin{tabular}{c|c|c}
        \hline
         & Manual annotation & Proposed method\\
        \hline\hline
     	    Total images
         		& 32845 & 32845 \\        
        \hline
     	    \multirow{3}{*}{\begin{tabular}{l}Time in total \\for objects \\and image\end{tabular}}
     	        & 10 [days] for recording images & 10 [days] for recording images \\
     	    	& 22.8 [hours] for annotation\tnote{1}    & 0 [min] for annotation \\
     		    & 0 [min] for paining           & Less than 5 [min] for painting \\
        \hline
     	    Time per image
         		& About 2.5 [min] / image & About 0 [min] / image \\
        \hline
     	    \multirow{4}{*}{\begin{tabular}{l}Money in total \\for objects \\and images\end{tabular}}
     	    	& \$ 200 for camera             & \$ 200 for camera \\
     		    & \$ 75 for LED                 & \$ 75 for LED \\
     		    &                               & \$ 25 for UV light \\
     		    & \$ 27590 for annotation\tnote{2}       & \$ 18 for fluorescent paint \\
        \hline
            Total cost & \$ 27865               & \$ 318 \\
     	    Money per image & \$ 0.85 / image               & \$ 0.0097 / image \\
        \hline
        \end{tabular}
        \begin{tablenotes}
            \item[1] \scriptsize {Average annotation time by AMT for this study is about 2.5 minutes per image.}
            \item[2] \scriptsize {Calculated from the minimum recommended semantic segmentation cost of \$ 0.84 per image in AMT.}
        \end{tablenotes}
    \endgroup
    \end{threeparttable}
\end{table}
\section{Conclusion}
\label{sec:conclusion}
In this paper, we proposed a method to create annotations automatically using \emph{invisible marker}, which is visible under UV light and invisible under regular light.
By switching between regular light and UV light at high speed, our system can create large datasets of dynamical changing deformable object manipulation in both controlled and uncontrolled environmental light conditions even with unfixed viewpoints.
The challenge of annotation gaps due to 1) the shift induced by the capture timing and 2) appearance change by \emph{invisible marker} can be absorbed sufficiently by the generalization capabilities of deep learning.
High accuracy of segmentation tasks is shown by multiple deep learning models such as FCN, U-Net, and SegNet trained on datasets created with our method.
We conclude that our proposed method can create large datasets accurately, quickly and inexpensively.

These datasets can widen the range of robotic tasks such as folding clothes, cooking, and biomedical applications. 
For future work, we would like to look into manipulation tasks with a robot such as cooking and laundry folding.
\section*{ACKNOWLEDGMENT} \small
The authors would like to thank Shunta Saito for helping to discuss and experiment, and Wilson Ko for proofreading and helping to write this article.
\bibliographystyle{IEEEtran} 
\bibliography{IEEEabrv,bibliography}
\end{document}